\documentclass[letterpaper]{article} 
\usepackage{aaai25}  
\usepackage{times}  
\usepackage{helvet}  
\usepackage{courier}  
\usepackage[hyphens]{url}  
\usepackage{graphicx} 
\usepackage{natbib}  
\usepackage{caption} 
\usepackage{algorithm}
\usepackage{algorithmic}

\usepackage{newfloat}
\usepackage{listings}
\DeclareCaptionStyle{ruled}{labelfont=normalfont,labelsep=colon,strut=off} 
\floatstyle{ruled}
\newfloat{listing}{tb}{lst}{}
\floatname{listing}{Listing}
\usepackage{amsmath,amssymb,amsfonts,bm}

\newcommand{\xbm}{{\bm x}}

\newcommand{\ybf}{\mathbf{y}}
\newcommand{\Xtid}{\Tilde{\mathbf{X}}}
\newcommand{\prompt}{\bm p}
\newcommand{\kbm}{{\bm k}}
\newcommand{\zbm}{{\bm z}}
\newcommand{\qbm}{{\bm q}}

\newcommand{\softmax}{\mathrm{softmax}}
\newcommand{\expo}{\mathrm{exp}}

\def\RR{\mathbb{R}}
\def\ie{{\em i.e.,~}}

\title{Adaptive Prompting for Continual Relation Extraction: \\ A Within-Task Variance Perspective}

\author{
    Minh Le \textsuperscript{\rm 1}\footnote{Work done at Hanoi University of Science and Technology.}\equalcontrib , 
    Tien Ngoc Luu \textsuperscript{\rm 2}\equalcontrib, 
    An Nguyen The \textsuperscript{\rm 3}\equalcontrib, 
    Thanh-Thien Le \textsuperscript{\rm 1$*$},
    Trang Nguyen \textsuperscript{\rm 1$*$}, \\
    Tung Thanh Nguyen \textsuperscript{\rm 4}, 
    Linh Ngo Van \textsuperscript{\rm 2}\thanks{Corresponding Author.}, Thien Huu Nguyen \textsuperscript{\rm 5}
}
\affiliations{
    \textsuperscript{\rm 1}VinAI Research\\ 
    \textsuperscript{\rm 2}Hanoi University of Science and Technology\\
    \textsuperscript{\rm 3}FPT Software AI Center\\
    \textsuperscript{\rm 4}Moreh Inc.\\
    \textsuperscript{\rm 5}University of Oregon, Eugene, Oregon, USA\\
    v.minhld12@vinai.io, ngoc.lt204595@sis.hust.edu.vn, annt68@fpt.com, v.thienlt3@vinai.io, v.trangnvt2@vinai.io, tung.nguyen@moreh.com.vn, linhnv@soict.hust.edu.vn, thien@cs.uoregon.edu

}

\usepackage{bibentry}

\begin{document}

\maketitle 

\begin{abstract}
To address catastrophic forgetting in Continual Relation Extraction (CRE), many current approaches rely on memory buffers to rehearse previously learned knowledge while acquiring new tasks. Recently, prompt-based methods have emerged as potent alternatives to rehearsal-based strategies, demonstrating strong empirical performance. However, upon analyzing existing prompt-based approaches for CRE, we identified several critical limitations, such as inaccurate prompt selection, inadequate mechanisms for mitigating forgetting in shared parameters, and suboptimal handling of cross-task and within-task variances. To overcome these challenges, we draw inspiration from the relationship between prefix-tuning and mixture of experts, proposing a novel approach that employs a prompt pool for each task, capturing variations within each task while enhancing cross-task variances. Furthermore, we incorporate a generative model to consolidate prior knowledge within shared parameters, eliminating the need for explicit data storage. Extensive experiments validate the efficacy of our approach, demonstrating superior performance over state-of-the-art prompt-based and rehearsal-free methods in continual relation extraction.
\end{abstract}

%

\section{Introduction}

Continual Relation Extraction (CRE) involves classifying semantic relationships between entities in text while adapting to an expanding set of relation types. Previous CRE approaches \cite{zhao-etal-2022-consistent, nguyen2023spectral,le2024continual} have successfully addressed the challenge of learning new relations without sacrificing accuracy on previously learned ones by employing memory-based techniques \cite{ shin2017continual, chaudhry2018efficient}. These methods utilize a rehearsal mechanism supported by a memory buffer, enabling the model to revisit and consolidate knowledge of prior relations while learning new tasks, thereby reducing catastrophic forgetting. Nonetheless, concerns regarding data storage and privacy have prompted the research community to investigate alternative strategies for CRE \cite{ke2022continual}.

To address these limitations, recent advances in Continual Learning (CL) have introduced innovative prompt-based methods \cite{wang2022learning, wang2022dualprompt, wang2023hide, tran2024leveraging}. Unlike memory-based approaches, these methods eliminate the need for rehearsal, focusing instead on the dynamic insertion of auxiliary parameters, known as prompts, during training. These prompts are adaptable to specific tasks, enabling continual learning without the necessity of data replay. However, our analysis identifies several inherent weaknesses in these prompt-based approaches. Firstly, they lack robust mechanisms to prevent forgetting in shared components, such as the shared Prompt Pool \cite{wang2022learning}, the task-agnostic General Prompt (G-Prompt) \cite{wang2022dualprompt}, or the shared MLP classifier, which can lead to potential performance degradation. Secondly, task-specific prompt-based approaches \cite{wang2022dualprompt, wang2023hide} are prone to inaccuracies in prompt selection, leading to a mismatch between training and testing prompts. Lastly, these methods demonstrate limited optimization in managing both cross-task and within-task variance. For instance, \citet{wang2022learning} employ a common prompt pool, where instances from different tasks may frequently share one or more prompts, thereby reducing cross-task variance. This issue becomes particularly prominent in CRE, where instances from different relation classes might frequently have very similar contexts, as shown in the example below:
\begin{itemize}
    \item \texttt{"[X] is a professor at [Z university]."}
    \item \texttt{"[X] is advised by a professor at [Z university]."}.
\end{itemize}

\begin{figure}
    \includegraphics[scale=0.29]{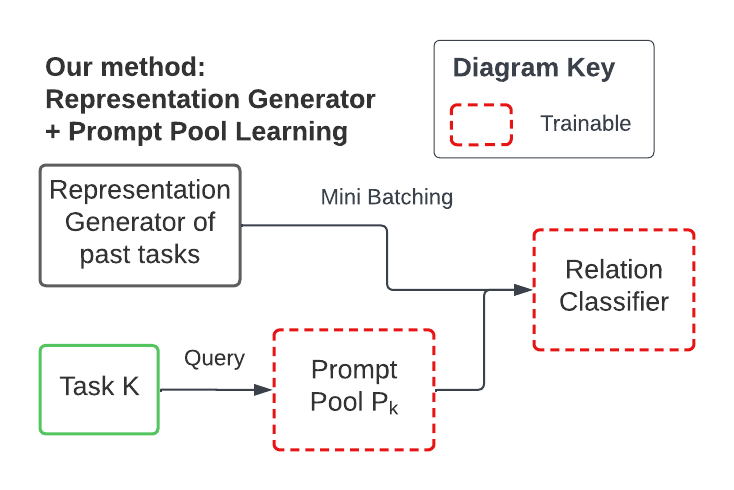} 
    \caption{\textbf{Overall framework  of WAVE-CRE.}
    To prevent information loss across tasks, we use a task-specific prompt pool $\mathbf{P}_t$ for each task and a representation generator to synthesize past-task information, strengthening the relation classifier's knowledge retention.
    }
    \label{fig:Our framework}
\end{figure}

In alignment with these approaches, \citet{le2024mixture} explores the relationship between Prefix-tuning \cite{li2021prefixtuning}, a widely used technique for implementing prompts, and Mixture of Experts (MoE) models \cite{Jacob_Jordan-1991, jordan1994hierarchical}. The study demonstrates that self-attention can be seen as embodying multiple MoE models, and that implementing prefix-tuning is analogous to adding new \emph{prefix} experts to these pre-trained MoE models to fine-tune their representations. Building on this foundation, we introduce a novel prompting method, \textbf{WAVE-CRE} (\underline{W}ithin-T\underline{a}sk \underline{V}ariance Awar\underline{e}ness for \underline{C}ontinual \underline{R}elation \underline{E}xtraction), designed to address the limitations highlighted earlier, as illustrated in Figure \ref{fig:Our framework}. During training, each task is allocated a dedicated Prompt Pool \cite{wang2022learning}, fostering cross-task divergence while capturing intra-task variations. To mitigate catastrophic forgetting during the shared parameter learning process between tasks, we employ a generative model that generates latent data representations for replay. Unlike generating natural language text, learning the underlying distribution and generating continuous latent representations is significantly more feasible. For prompt pool selection, a separate generative model generates uninstructed representations, which are then utilized to train the task predictor. Extensive experiments demonstrate that our method surpasses state-of-the-art prompt-based and rehearsal-free baselines. Therefore, our contributions are as follows: 
\begin{itemize}
    \item  We reveal limitations of current prompt-based approaches, including inaccurate prompt selection, inadequate strategies for mitigating forgetting in shared parameters, and suboptimal handling of cross-task and within-task variances.
    \item To overcome these challenges, we propose a novel prompting method, WAVE-CRE, which leverages task-specific prompt pools and generative models for latent representation.
    \item Our extensive experimental evaluation demonstrates that WAVE-CRE significantly outperforms state-of-the-art prompt-based and rehearsal-free baselines.
\end{itemize}

\section{Background} \label{section:background}

\subsection{Continual Relation Extraction}
\label{problem}
Continual Relation Extraction (CRE) is a subfield of continual learning \cite{hai2024continual,phan2022reducing,linhp} and continual information extraction \cite{le2024sharpseq,dao2024lifelong,tran2024preserving}. It involves training a model sequentially on a series of tasks $\{\mathcal{T}_1, \mathcal{T}_2, \ldots, \mathcal{T}_T\}$, with each task $\mathcal{T}_t$ associated with a training dataset $\mathcal{D}_t$ and a corresponding set of relations $\mathcal{R}_t$. Similar to traditional supervised classification frameworks \cite{ji-etal-2020-span}, each task $\mathcal{T}_t$ consists of $\mathcal{N}_t$ labeled samples, $\mathcal{D}_t =  \{(\boldsymbol{x}^t_i, y^t_i)\}_{i=1}^{\mathcal{N}_t}$, where $\boldsymbol{x}^t_i$ denotes the input data, and $y^t_i$ corresponds to a label from the relation set $\mathcal{R}_t$. The primary objective in CRE is to train a model that can effectively learn from new tasks while preserving its performance on previously acquired tasks. Upon completing the $t$-th task, the model should be able to accurately identify the relation of an entity pair from the cumulative relation set $\hat{\mathcal{R}}_t=\bigcup_{i=1}^t \mathcal{R}_i$. Most existing methods in CRE rely on the use of a memory buffer to store samples of previously encountered relations, which raises significant concerns regarding memory and privacy. 


\subsection{Mixture of Experts} \label{section:background_moe}
 
An MoE model consists of $N$ expert networks, denoted by $f_i: \mathbb{R}^D \rightarrow \mathbb{R}^{D_v}$ for $i = 1, \dots, N$, and a gating function, $G: \mathbb{R}^D \rightarrow \mathbb{R}^{N}$, which dynamically determines the contribution of each expert for a given input $\xbm$. The gating function is based on learned score functions, $s_i: \mathbb{R}^D \rightarrow \mathbb{R}$, associated with each expert, resulting in the following formulation:
\begin{align}
    \ybf &= \sum_{j=1}^N G(\xbm)_j \cdot f_j(\xbm) = \sum_{j=1}^N \frac{\exp\left(s_j(\xbm)\right)}{\sum_{\ell=1}^N\exp\left(s_\ell(\xbm)\right)} \cdot f_j(\xbm),
\end{align}
where $G(\xbm) = \softmax(s_1(\xbm),\ldots,s_N(\xbm))$. \citet{Quoc-conf-2017} introduced Sparse Mixture of Experts (SMoE) architecture as an efficient method to scale up MoE models. This is achieved by utilizing a sparse gating function $\mathrm{TopK}$, which selects only the $K$ experts with the largest affinity scores $s_j(\xbm)$. The $\mathrm{TopK}$ function is defined as:
\begin{align*}
\nonumber & \mathrm{TopK}\left(\bm v, K\right)_i \\ 
& =\left\{\begin{array}{cc}
\bm v_i, & \text { if } \bm v_i \text { is in the } K \text { largest elements of } \bm v \\
-\infty, & \text { otherwise. }
\end{array}\right.
\end{align*}
Subsequently, the selected experts independently calculate their outputs, and these are linearly combined using their corresponding affinity scores to produce the final prediction:
\begin{align}
\ybf=\sum_{j=1}^N \softmax\left(\mathrm{TopK}\left(s(\xbm), K\right)\right)_j \cdot  f_j(\xbm),
\end{align}
where $s(\xbm) = (s_1(\xbm),\dots, s_N(\xbm))$. MoE has gained significant attention for its flexibility and adaptability in fields like large language models \cite{glam, brainformers}, computer vision \cite{visionmoe}, and multi-task learning \cite{mmoe}.

\subsection{Prompt-based Methods}
\label{sec:prompt-based-cl}

Recently, parameter-efficient fine-tuning techniques like Prompt-tuning \cite{lester2021power} and Prefix-tuning \cite{liu2022ptuning, le2024revisiting} have become prominent for fine-tuning pre-trained models on downstream tasks. This study focuses on prefix-tuning for prompt implementation, where prompts are passed to several Multi-head Self-attention (MSA) layers in the pre-trained transformer encoder. Let $\mathbf{X} = [\xbm_1,\dots,\xbm_N]^\top \in \mathbb{R}^{N \times D}$ represent the input matrix, where $\xbm_i$ is the embedding of the $i$-th token, and $D$ is the embedding dimension. The output of an MSA layer is:
\begin{align}
    &\mathrm{MSA}(\mathbf{X}) = \mathrm{Concat}(h_1, ..., h_{m})W_O,  \\
    &h_i = \mathrm{Attention}(\mathbf{X}W_i^Q, \mathbf{X}W_i^K, \mathbf{X}W_i^V), \label{eq:msa}
\end{align}
for $i = 1, \dots, m$, where $W_O$ , $W_i^Q$, $W_i^K$, and $W_i^V$ are projection matrices, and $m$ is the number of heads. In prefix-tuning, a prompt $P \in \mathbb{R}^{L_p \times D}$ of length $L_p$ is divided into $P_k, P_v \in \mathbb{R}^{\frac{L_p}{2} \times D}$. Each head $h_i$ calculation is modified as:
\begin{align}
    \hat{h}_i = \mathrm{Attention}(\mathbf{X}W_i^Q, [P_k; \mathbf{X}]W_i^K, [P_v; \mathbf{X}]W_i^V), \label{eq:prefix_tuning}
\end{align}
where $[\cdot; \cdot]$ denotes the concatenation operation along the sequence length dimension. 

Recent research by \citet{le2024mixture} has demonstrated that self-attention can be interpreted as a specialized architecture comprising multiple MoE models. The study further suggests that prefix-tuning functions as a mechanism to introduce new experts into these MoE models. Specifically, consider the $l$-th head in the MSA layer, with output $h_l = [h_{l, 1}, \dots, h_{l, N}]^\top \in \RR^{N \times D_v}$. Let $\Xtid = [\xbm_1^\top,\dots,\xbm_N^\top]^\top \in \mathbb{R}^{N \cdot D}$ represent the concatenation of all input token embeddings. We define $N$ experts $f_j: \RR^{N \cdot D} \rightarrow \RR^{D_v}$ and $N$ gating functions $G_i: \RR^{N \cdot D} \rightarrow \RR^N$ with input $\Xtid$ as follows:
\begin{align}
    f_j(\Xtid) &= {W_l^V}^\top E_{j} \Xtid = {W_l^V}^\top \xbm_j, \label{eq:pre-trained_experts} \\
    G_i(\Xtid) &= \softmax(s_{i, 1}(\Xtid), \dots, s_{i, N}(\Xtid)), \nonumber \\
    s_{i,j}(\Xtid) &= \frac{\Xtid^{\top} E_{i}^{\top} W_l^Q  {W_l^K}^\top E_{j} \Xtid}{\sqrt{D_v}}, \label{eq:pre-trained_scores}
\end{align}
for $i, j = 1, \dots, N$, where $E_{i} \in \mathbb{R}^{D \times N \cdot D}$ are matrices such that $E_{i} \Xtid = \xbm_{i}$, and $D_v = \frac{D}{m}$ is the key dimension. Then, from equation~\eqref{eq:msa}, the output of the $l$-th head can be expressed as:
\begin{align}
    h_{l, i} &= \sum_{j=1}^N G_i(\Xtid)_j \cdot f_j(\Xtid) \nonumber \nonumber \\ 
    &= \sum_{j=1}^N \frac{\exp\left(s_{i, j}(\Xtid)\right)}{\sum_{\ell=1}^N\exp\left(s_{i, \ell}(\Xtid)\right)} \cdot f_j(\Xtid), \label{eq:attn_moe}
\end{align}
for $i = 1, \dots, N$. From equation~\eqref{eq:attn_moe}, we observe that each head $h_l$ in the MSA layer includes $N$ MoE models $h_{l, 1}, \dots, h_{l, N}$. This structure is similar to the Multi-gate Mixture of Experts \cite{mmoe}, where multiple MoE models leverage the same set of expert networks but employ independent gating functions. Extending this concept, \citet{le2024mixture} proposes that prefix-tuning can be viewed as a method for incorporating new experts into these MoE models. New prefix experts and score functions can be defined as:
\begin{align}
    f_{N + j}(\Xtid) &= {W_l^V}^\top \prompt^v_j, \label{eq:prompt_expert} \\
    s_{i, N + j}(\Xtid) &= \frac{\Xtid^\top E_{i}^{\top} W_l^Q  {W_l^K}^\top \prompt^k_j}{\sqrt{D_v}} , \label{eq:prompt_score}
\end{align}
for $i = 1,\dots,N$ and $j=1,\dots,L$, where $P_k = [\prompt^k_1, \dots, \prompt^k_L]^\top$, $P_v = [\prompt^v_1, \dots, \prompt^v_L]^\top$, and $L = \frac{L_p}{2}$. From equation~\eqref{eq:prefix_tuning}, the output of the $l$-th head can be written as $\hat{h}_l = [\hat{h}_{l, 1}, \dots, \hat{h}_{l, N}]^\top \in \RR^{N \times D_v}$, where
\begin{align}
\hat{h}_{l, i}
    &= \sum_{j = 1}^N  
        \frac{\exp(s_{i, j}(\Xtid))}
        {
            \sum_{k = 1}^{N + L} \exp(s_{i, k}(\Xtid))
        } f_j(\Xtid) \nonumber \\
    &+ \sum_{j' = 1}^L  
    \frac{\exp(s_{i, N + j'}(\Xtid))}
    {
        \sum_{k = 1}^{N + L} \exp(s_{i, k}(\Xtid))
    } f_{N + j'}(\Xtid), \label{eq:prefix_moe}
\end{align}
for $i = 1, \dots, N$. These new experts, $f_{N + 1}, \dots, f_{N + L}$, collaborate with the pre-trained experts $f_1, \dots, f_N$ to adapt the model for downstream tasks. Several recent methods have effectively integrated prompting techniques with pre-trained transformer encoders, yielding notable results as demonstrated by L2P \cite{wang2022learning}, DualPrompt \cite{wang2022dualprompt}, and HiDe-Prompt \cite{wang2023hide}. 

\section{Methodology} \label{sec:method}

\begin{figure*}
    \includegraphics[scale=0.28]{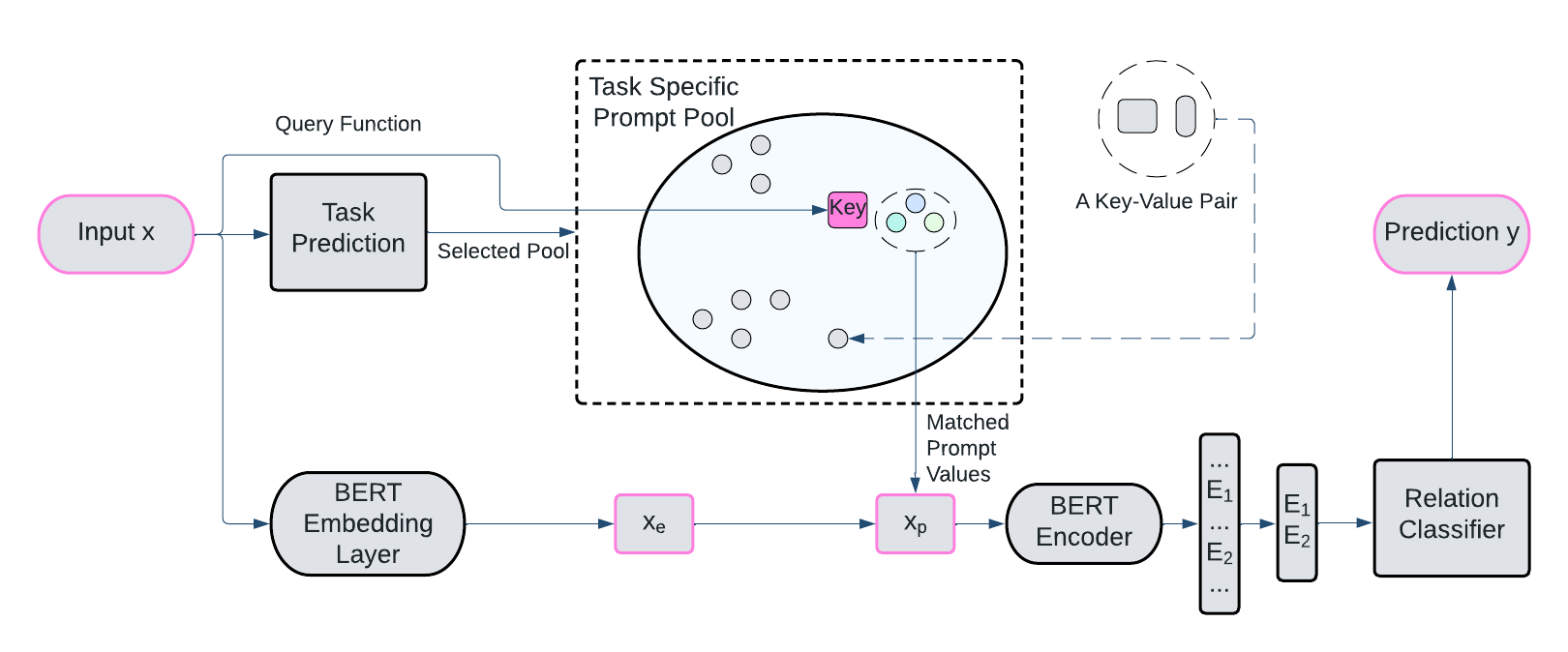}
    \captionsetup{justification=justified, singlelinecheck=false}
    \captionsetup{skip=-7.5pt}
    \caption{\textbf{Data Flow Diagram}: Initially, the task predictor predicts the task identity of the input $\boldsymbol{x}$, enabling the selection of the corresponding prompt pool. Subsequently, the input $\boldsymbol{x}$ queries this prompt pool to identify prompts whose corresponding keys are closest to the $query \ q(\xbm)$. The chosen prompt is then prepended to the embedded input $\boldsymbol{x}_e$, creating the prompted input $\boldsymbol{x}_p$. The combined $\boldsymbol{x}_p$ is fed into the BERT Encoder, where the two embeddings corresponding to the positions of the entities $E_1$ and $E_2$ are concatenated. Finally, the resulting concatenated embedding is passed to the relation classifier, which predicts the relation label $y$ of the input $\boldsymbol{x}$.}
    \label{fig:data-flow-diagram}
\end{figure*}

Our overall approach, depicted in Figure~\ref{fig:Our framework}, involves three main stages: (1) Prompt pool learning for a new task; (2) Generative models; and (3) Training the task predictor and relation classifier.

\subsection{Task-specific Prompt Pool} \label{section:task-specific-prompt-pool}

In our approach to CRE, we adopt a frozen BERT model \cite{devlin2018bert} as the pre-trained transformer encoder, maintaining consistency with prior studies \cite{xia2023enhancing, zhao2023improving}.

Previous approaches, such as HiDe-Prompt \cite{wang2023hide}, utilize a single prompt per task. This strategy can be compared to utilizing a fixed set of experts for every instance within a given task. However, as detailed in equation~\eqref{eq:pre-trained_experts} and equation~\eqref{eq:prompt_expert}, the prefix experts encoded in these prompts are considerably simpler, functioning as offset vectors  rather than pre-trained experts, which are linear functions of the input. This inherent simplicity implies that a fixed set of prefix experts may lack the necessary flexibility to effectively capture the full range of task variations. To address this limitation, we extend the concept of the prompt pool introduced in L2P \cite{wang2022learning} by proposing a task-specific prompt pool. For each task $t$, we introduce a  prompt pool $\mathbf{P}_t$:
\begin{align}
\mathbf{P}_t = \{(\kbm_1^{(t)}, P_1^{(t)}), (\kbm_2^{(t)}, P_2^{(t)}), ..., (\kbm_M^{(t)}, P_M^{(t)})\} \label{eq:prompt_pool},
\end{align}
where $M$ represents the number of prompts. Each prompt $P_i^{(t)}$ is associated with a learnable key $\kbm_i^{(t)} \in \mathbb{R}^{D}$. To facilitate prompt selection, we adopt the same key-query mechanism described in L2P. Specifically, given an input sentence $\xbm$, it is first encoded using a pre-trained BERT model to generate a query vector $q(\xbm)$. A scoring function $\gamma: \mathbb{R}^D \times \mathbb{R}^D \rightarrow \mathbb{R}$ then evaluates the match between the query vector and each prompt key (e.g., using cosine similarity). The top $K$ most relevant prompts are selected by optimizing the following objective:
\begin{equation}  \label{equation:K_x}
\boldsymbol{K}_{\boldsymbol{x}} = \underset{S \subseteq \{1, \dots, M\}: |S| = K}{\rm argmin} \sum_{s \in S}\gamma(q(\boldsymbol{x}), \boldsymbol{k}_{s}^{(t)}),
\end{equation}
where $\boldsymbol{K}_{\boldsymbol{x}}$ denotes a subset of the top-$K$ keys that are specifically chosen for $\boldsymbol{x}$. 

We strategically set the number of experts within a prompt to one, with $L_p = 2$, resulting in $L = \frac{L_p}{2} = 1$. Using prompts of a larger length would be akin to incorporating more experts per prompt, all of which would share a common prompt key within the prompt pool. However, our approach not only reduces memory costs but also enhances flexibility in selecting experts during training and testing. This configuration allows each expert to adapt to different inputs, providing a more versatile assignment compared to methods that use multiple experts per prompt.

Our proposed architecture enables the assignment of different sets of prompts or experts to specific regions of the input data, guided by the contextual query feature $q(\xbm)$. This design allows each prompt within the pool to selectively focus on the relevant information and patterns necessary for optimal performance in distinct areas of the input domain. As a result, the model is capable of capturing within-task variations effectively. Furthermore, by utilizing a task-specific prompt pool, we reduce the need for parameter sharing, which helps maximize cross-task variance.

\paragraph{Relationship with Sparse Mixture of Experts.} Our proposed task-specific prompt pool shares certain similarities with the SMoE architecture in Section~\ref{section:background_moe}. Specifically, from equation~\eqref{eq:prompt_pool}, we denote the experts encoded in the prompt pool $\mathbf{P}_t$ as $f_{N + 1}^{(t)},\dots,f_{N + M}^{(t)}$. Unlike the pre-trained experts $f_1,\dots,f_N$, which are selected by default, we employ sparse selection exclusively for these newly introduced prefix experts. As illustrated in equation~\eqref{eq:prefix_moe}, each head in the MSA layer when applying prefix-tuning encompasses $N$ MoE models $\hat{h}_{l, 1}, \dots, \hat{h}_{l, N}$. The standard approach involves applying the $\mathrm{TopK}$ function to each of these $N$ models individually, necessitating the computation of all $N \times M$ score functions $s_{i, N + j}(\Xtid)$ for $i = 1, \dots, N$ and $j = 1, \dots, M$. This results in a distinct set of prefix experts selected for each model. Conversely, our strategy leverages the same set of $K$ new experts across all $N$ MoE models using auxiliary score functions defined as:
\begin{align}
    \hat{s}_{i, N + j}(\Xtid) = \gamma(q(\boldsymbol{x}), \boldsymbol{k}_{j}^{(t)}),
\end{align}
where $i=1,\dots,N$ and $j=1,\dots,M$. This approach only requires the computation of $M$ score functions, as the computation of $\hat{s}_{i, N + j}(\Xtid)$ only depends solely on $\boldsymbol{k}_{j}^{(t)}$, thereby enabling the efficient and effective selection of $K$ experts from the prompt pool. Although the computation of $q(\xbm)$ might appear to be an added expense, this value is already calculated for the task predictor in Section~\ref{section:classifiers} during both the training and testing phases. Therefore, the computation of $q(\xbm)$ can be reused, incurring no additional cost in the prompt selection process.

\paragraph{Optimization Objective.} For each new task $\mathcal{T}_t$, a new prompt pool $\mathbf{P}_t$ is created. During each training step, following the aforementioned strategy, $K$ prompts are selected, and the corresponding prompted embedding feature, denoted as $\boldsymbol{x}_p$, is inputted to the pre-trained transformer encoder $f_r$ and the final classifier $g_\phi$, which is parameterized by $\phi$. The objective can be summarized as follows:
\begin{equation} \label{eq:prompt-pool-learning-loss}
\min_{\mathbf{P}_t, \phi}\mathcal{L}(g_{\phi}(f_r(\boldsymbol{x}_p)), y) + \lambda\sum_{s_i \in \boldsymbol{K}_{\boldsymbol{x}}}\gamma(q(\boldsymbol{x}), \boldsymbol{k}_{s_i}^{(t)}),
\end{equation}
where $\boldsymbol{K}_{\boldsymbol{x}}$ is obtained with equation~\eqref{equation:K_x}, $\gamma$ denotes the cosine similarity function, and $\lambda$ is a hyperparameter. The first term in equation~\eqref{eq:prompt-pool-learning-loss} employs the classification loss, while the second term  minimizes the distance between prompt keys and the corresponding query features. 
Note that only the prompt parameters in $\mathbf{P}_t$ and the final classifier $g_{\phi}$ are learned during training task $t$. The pre-trained BERT model and previous pools $\mathbf{P}_1, \dots, \mathbf{P}_{t - 1}$ remain frozen.

\subsection{Generative Models for Relation Representation} 
\label{section:GMM}

To effectively retain knowledge acquired from prior tasks, we utilize a generative model that captures the distributions of observed relations, enabling the replay of relation samples. For each relation $r \in \hat{\mathcal{R}}_t$, we maintain a distribution $\mathbf{G}^r_z \sim \mathcal{N}(\boldsymbol{\mu}^r_z,\,\Sigma^r_z)$. This distribution is obtained by fitting a Gaussian distribution to the set $\mathcal{D}^r_z = \{\zbm^r = f_r(\xbm^r_p)\}$, which represents the prompted representation of the input $\xbm^r$ for relation $r$:
\begin{align} \label{eq:gauss_latent}
    \boldsymbol{\mu}^r_z = \sum_{\zbm^r} \frac{\zbm^r}{| \mathcal{D}^r_z |}, \
    \Sigma^r_z = \sum_{\zbm^r} 
        \frac{(\zbm^r - \boldsymbol{\mu}^r_z)(\zbm^r - \boldsymbol{\mu}^r_z)^\top}
        {| \mathcal{D}^r_z |}.
\end{align}
Similarly, the distribution of the query corresponding to each relation denoted as $\mathcal{D}^r_q = \{\qbm^r = q(\xbm^r)\}$, is stored as $\mathbf{G}^r_q \sim \mathcal{N}(\boldsymbol{\mu}^r_q,\,\Sigma^r_q)$:
\begin{align} \label{eq:gauss_task_prediction}
    \boldsymbol{\mu}^r_q = \sum_{\qbm^r} \frac{\qbm^r}{| \mathcal{D}^r_q |}, \
    \Sigma^r_q = \sum_{\qbm^r} 
        \frac{(\qbm^r - \boldsymbol{\mu}^r_q)(\qbm^r - \boldsymbol{\mu}^r_q)^\top}
        {| \mathcal{D}^r_q |}.
\end{align}
This approach ensures that we have a generative model for each relation, capable of reconstructing the corresponding prompted and query representations for all previously observed relations without storing any instance-specific data. The choice of a Gaussian distribution is motivated by its memory efficiency, as it requires storing only the mean vectors and covariance matrices, thereby minimizing the overall memory footprint. Future works may explore alternative generative models.

\subsection{Task Predictor and Relation Classifier} \label{section:classifiers}

Each task is associated with its own prompt pool, designed to adapt and learn from the samples specific to the task. However, at test time, it becomes crucial to identify which prompt pool corresponds to a new, unseen sample. To tackle this challenge, we introduce a task predictor, denoted as $\hat{g}_\psi$. This predictor is a feed-forward MLP with an output dimension matching the total number of relations encountered thus far (\ie $|\hat{\mathcal{R}}_t|$). Once trained, $\hat{g}_\psi$ is capable of predicting the task identity, thereby facilitating the selection of the appropriate prompt pool during testing.

To train the task predictor, we utilize $\mathbf{G}^r_q$ described in Section \ref{section:GMM}  to generate a representation set $\qbm^r$ for each relation $r \in \hat{\mathcal{R}}_t$. The task predictor is trained with the cross-entropy loss function defined as:
\begin{align} \label{eq:loss_task_predictor}
    \mathcal{L}(\psi) = \sum_{r \in \hat{\mathcal{R}}_t} \sum_{\qbm \sim \mathbf{G}^r_q}
        -\mathrm{log} \frac
        {\expo(\hat{g}_\psi(\qbm)[r])}
        {\sum_{r' \in \hat{\mathcal{R}}_t} \expo(\hat{g}_\psi(\qbm)[r'])}.
\end{align}
While our approach shares similarities with HiDe-Prompt \cite{wang2023hide} in utilizing an additional MLP head, a key distinction lies in how relations are treated. HiDe-Prompt categorizes all relations within a task as a single class in the cross-entropy loss. This strategy can be suboptimal, as the resulting classes may lack semantic significance and their meaning can depend on the sequence in which tasks are presented during training.

In a manner analogous to the training of the task predictor, we train the relation classifier $g_\phi$ using $\mathbf{G}^r_z$ with the same cross-entropy loss function defined as follows:
\begin{align} \label{eq:loss_relation_classifier}
    \mathcal{L}(\phi) = \sum_{r \in \hat{\mathcal{R}}_t} \sum_{\zbm \sim \mathbf{G}^r_z}
        -\mathrm{log} \frac
        {\expo(g_\phi(\zbm)[r])}
        {\sum_{r' \in \hat{\mathcal{R}}_t} \expo(g_\phi(\zbm)[r'])}.
\end{align}
This approach helps to mitigate catastrophic forgetting in the shared classification head without requiring the storage of samples from previous relations. For a detailed overview of the training process, please refer to Algorithm~\ref{alg:cap}. The data flow diagram illustrating the inference process is provided in Figure~\ref{fig:data-flow-diagram}.

\begin{algorithm}[tb]
\caption{$\mathcal{T}_t$ training process} \label{alg:cap}
\label{alg:algorithm}
\textbf{Input}: Training $t$-th dataset $\mathcal{D}_t$, current relation set $\mathcal{R}_t$\\
\textbf{Output}: Prompt pool $\mathbf{P}_t$, task predictor $\psi$, and relation classifier $\phi$
\begin{algorithmic}[1] 
\STATE Randomly initialize $\mathbf{P}_t$
\FOR{$e_{id} \gets 1$ \textbf{to} $training\_epoch$}
    \FOR{batch $\boldsymbol{x}_{B} \in \mathcal{D}_t$}
        \STATE Update $\mathbf{P}_t$ and $g_{\phi}$ on $\boldsymbol{x}_{B}$ via equation~\eqref{eq:prompt-pool-learning-loss}
    \ENDFOR
\ENDFOR
\STATE Update $\hat{\mathcal{R}}_{t} \gets \hat{\mathcal{R}}_{t-1} \cup \mathcal{R}_t$

\FOR{each $r \in \mathcal{R}_t$}
    \STATE $\mathcal{D}^r_q \gets \emptyset, \ \mathcal{D}^r_z \gets \emptyset$
    \FOR{batch $\boldsymbol{x}_{B} \in \mathcal{D}_t^r$}
        \STATE Update $\mathcal{D}^r_q, \mathcal{D}^r_z$
    \ENDFOR
    \STATE Fit $\mathbf{G}_z^r$ to $\mathcal{D}^r_z$ via equation~\eqref{eq:gauss_latent}
    \STATE Fit $\mathbf{G}_q^r$ to $\mathcal{D}^r_q$ via equation~\eqref{eq:gauss_task_prediction}
\ENDFOR
\STATE Train the task predictor $\psi$ via equation~\eqref{eq:loss_task_predictor}
\STATE Train the relation classifier $\phi$ via equation~\eqref{eq:loss_relation_classifier}
\STATE \textbf{return} $\mathbf{P}_{t},  \phi, \psi$
\end{algorithmic}
\end{algorithm}

\section{Experiments}
\label{sec:experiment}

\begin{table*} [t]
\small
\centering
    \begin{tabular}{l | c c c c c c c c c c} 
    \hline
    \multicolumn{11}{c}{\textbf{FewRel}} \\
    \hline
    Model & $\mathcal{T}_{1}$ & $\mathcal{T}_{2}$ & $\mathcal{T}_{3}$ & $\mathcal{T}_{4}$ & $\mathcal{T}_{5}$ & $\mathcal{T}_{6}$ & $\mathcal{T}_{7}$ & $\mathcal{T}_{8}$ & $\mathcal{T}_{9}$ & $\mathcal{T}_{10}$ \\
    \hline
    EA-EMR & 89.0 & 69.0 & 59.1 & 54.2 & 47.8 & 46.1 & 43.1 & 40.7 & 38.6 & 35.2 \\
    RP-CRE & 97.9 & 92.7 & 91.6 & 89.2 & 88.4 & 86.8 & 85.1 & 84.1 & 82.2 & 81.5 \\
    CRL & 98.2 & 94.6 & 92.5 & 90.5 & 89.4 & 87.9 & 86.9 & 85.6 & 84.5 & 83.1\\
    CRE-DAS & 98.1 & \underline{95.8} & \underline{93.6} & 91.9 & \underline{91.1} & 89.4 & 88.1 & 86.9 & 85.6 & 84.2 \\
    CDec+ACA & \underline{98.4} & 95.4 & 93.2 & \underline{92.1} & 91.0 & \underline{89.7} & \underline{88.3} & \underline{87.4} & \underline{86.4} & \underline{84.8} \\
    \hline
    L2P & 97.4	& 90.8 & 83.6 & 76.5 & 68.9 & 64.1 & 61.0 & 57.4 & 50.1 & 44.6 \\
    EPI & \textbf{98.3}	&89.9	&84.0	&79.9	&76.5	&73.1	&70.1	&67.0	&64.5	&61.8 \\
    HiDe-Prompt & 95.5	&89.4	&86.0	&85.7	&87.8	&84.2	&75.9	&75.1	&70.3	&67.2 \\
    \hline
    WAVE-CRE & 97.9 & \textbf{95.5} & \textbf{93.6} & \textbf{92.4} &	\textbf{91.1} & \textbf{90.2} & \textbf{88.7} & \textbf{87.6} & \textbf{86.5} & \textbf{85.0} \\
    \hline
    \end{tabular}

    \begin{tabular}{l | c c c c c c c c c c} 
    \hline
    \multicolumn{11}{c}{\textbf{TACRED}} \\
    \hline
    Model & $\mathcal{T}_{1}$ & $\mathcal{T}_{2}$ & $\mathcal{T}_{3}$ & $\mathcal{T}_{4}$ & $\mathcal{T}_{5}$ & $\mathcal{T}_{6}$ & $\mathcal{T}_{7}$ & $\mathcal{T}_{8}$ & $\mathcal{T}_{9}$ & $\mathcal{T}_{10}$ \\
    \hline
    EA-EMR & 47.5 & 40.1 & 38.3 & 29.9 & 24 & 27.3 & 26.9 & 25.8 & 22.9 & 19.8 \\
    RP-CRE & 97.6 & 90.6 & 86.1 & 82.4 & 79.8 & 77.2 & 75.1 & 73.7 & 72.4 & 72.4 \\
    CRL & \underline{97.7} & 93.2 & 89.8 & 84.7 & 84.1 & 81.3 & 80.2 & 79.1 & 79.0   & 78.0\\
    CRE-DAS & \underline{97.7} & \underline{94.3} & \underline{92.3} & \underline{88.4} & \underline{86.6} & \underline{84.5} & \underline{82.2} & \underline{81.1} & \underline{80.1} & \underline{79.1} \\
    CDec+ACA & \underline{97.7} & 92.8 & 91.0 & 86.7 & 85.2 & 82.9 & 80.8 & 80.2 & 78.8 & 78.6 \\
    \hline
    L2P & 96.9 & 88.2 & 73.8 & 68.6 & 66.3 & 63.1 & 60.4 & 59.1 & 56.8 & 54.8 \\
    EPI &97.5	&90.7	&82.7	&76.7	&74.0	&72.3	&68.2	&66.5	&65.1	&63.4 \\
    HiDe-Prompt &97.3	&92.8	&86.2	&82.6	&80.6	&80.4	&75.8	&73.7	&72.9	&72.6 \\
    \hline
    WAVE-CRE & \textbf{98.4} & \textbf{94.3} & \textbf{91.6} & \textbf{87.8} & \textbf{85.7} & \textbf{83.5} & \textbf{81.3} & \textbf{80.4} & \textbf{79.5} & \textbf{78.7} \\
    \hline
    \end{tabular}
\caption{Average accuracy (\%) of all methods across learning stages for FewRel and TACRED dataset. The best accuracy scores under the rehearsal-free and rehearsal-based setting are in \textbf{bold} and \underline{underlined}, respectively.}
\label{tab:fewrel-res}
\end{table*}
\begin{table*} [t]
\small
\centering
    \begin{tabular}{l | c c c c c c c c c c} 
    \hline
    \multicolumn{11}{c}{\textbf{Task-Incremental Learning - TACRED}} \\
    \hline
    Model & $\mathcal{T}_{1}$ & $\mathcal{T}_{2}$ & $\mathcal{T}_{3}$ & $\mathcal{T}_{4}$ & $\mathcal{T}_{5}$ & $\mathcal{T}_{6}$ & $\mathcal{T}_{7}$ & $\mathcal{T}_{8}$ & $\mathcal{T}_{9}$ & $\mathcal{T}_{10}$ \\
    \hline
    WAVE-CRE & \textbf{98.4} & \textbf{95.5} & \textbf{94.2} &	\textbf{94.1} & \textbf{92.7} & \textbf{89.9} & \textbf{88.3} &	\textbf{87.6} &	\textbf{86.5} & \textbf{85.2} \\
    w/o Prompt Pool & 96.8 & 94.0 & 92.9 &	91.5 & 	90.6 &	88.0 &	86.1 &	84.9 &	84.4 &	83.4 \\
    \hline
    \end{tabular}
\caption{Detailed analysis of WAVE-CRE with task-specific prompt pool in the task-incremental learning scenario of TACRED.  We report the average accuracy across different stages. The best accuracy scores are in \textbf{bold}.}
\label{tab:ablation_prompt_pool}
\end{table*}

\begin{table*} [ht]
\small
\centering
    \begin{tabular}{c c | c c c c c c c c c c} 
    \hline
    \multicolumn{12}{c}{\textbf{Task Incremental Learning - TACRED}} \\
    \hline
    $L$ & $K$ & $\mathcal{T}_{1}$ & $\mathcal{T}_{2}$ & $\mathcal{T}_{3}$ & $\mathcal{T}_{4}$ & $\mathcal{T}_{5}$ & $\mathcal{T}_{6}$ & $\mathcal{T}_{7}$ & $\mathcal{T}_{8}$ & $\mathcal{T}_{9}$ & $\mathcal{T}_{10}$ \\
    \hline
    8 & 1 & 97.2 & 94.8	 &  93.1 & 92.3	& 90.5 & 87.8	&84.6	&84.1	& 83.4	& 84.2 \\
    4 & 2 & 97.7 &	95.5 &	94.9 & 91.2 & 90.7 & 88.0 & 86.5 &	86.3 &	85.3 &	84.1  \\
    2 & 4 & 96.5	& 95.1	&93.2	&92.1	&91.2	&88.9	&85.3	&85.2	& 84.6	& 84.0 \\
    1 & 8  & \textbf{98.4} & \textbf{95.5} & \textbf{94.2} &	\textbf{94.1} & \textbf{92.7} & \textbf{89.9} & \textbf{88.3} &	\textbf{87.6} &	\textbf{86.5} & \textbf{85.2} \\
    \hline
    \end{tabular}
\caption{Detailed analysis of the impact of the number of experts within a prompt. We report the average accuracy across different stages on TACRED in the task incremental learning scenario. The best accuracy scores are in \textbf{bold}.}  
\label{tab:ablation_prompt_length}
\end{table*}
\begin{table*} [t]
\small
\centering
    \begin{tabular}{l | c c c c c c c c c c} 
    \hline
    \multicolumn{11}{c}{\textbf{FewRel}} \\
    \hline
    Model & $\mathcal{T}_{1}$ & $\mathcal{T}_{2}$ & $\mathcal{T}_{3}$ & $\mathcal{T}_{4}$ & $\mathcal{T}_{5}$ & $\mathcal{T}_{6}$ & $\mathcal{T}_{7}$ & $\mathcal{T}_{8}$ & $\mathcal{T}_{9}$ & $\mathcal{T}_{10}$ \\
    \hline
    EPI &64.5	&62.8	&64.4	&64.9	&64.5	&64.4	&64.4	&59.1	&61.1	&56.6 \\
    HiDe-Prompt &76.7	&81.7	&80.6	&80.6	&80.2	&78.9	&77.8	&83.1	&80.3	&81.0 \\

    \hline
    WAVE-CRE & \textbf{88.5} & \textbf{86.2} & \textbf{86.9} & \textbf{86.4} & \textbf{85.2} & \textbf{86.9} & \textbf{87.7} & \textbf{86.0} & \textbf{85.4} & \textbf{82.5} \\
    \hline
    \end{tabular}
    \begin{tabular}{l | c c c c c c c c c c} 
    \hline
    \multicolumn{11}{c}{\textbf{TACRED}} \\
    \hline
    Model & $\mathcal{T}_{1}$ & $\mathcal{T}_{2}$ & $\mathcal{T}_{3}$ & $\mathcal{T}_{4}$ & $\mathcal{T}_{5}$ & $\mathcal{T}_{6}$ & $\mathcal{T}_{7}$ & $\mathcal{T}_{8}$ & $\mathcal{T}_{9}$ & $\mathcal{T}_{10}$ \\
    \hline
    EPI &64.2	&58.8	&66.2	&55.3	&64.0	&62.1	&67.8	&62.0	&62.4	&62.5 \\
    HiDe-Prompt &70.2	&72.0	&73.8	&75.9	&68.6	&\textbf{79.8}	&75.2	& \textbf{71.2}	&68.5	&64.9 \\

    \hline
    WAVE-CRE & \textbf{83.8} & \textbf{79.5} & \textbf{84.2} & \textbf{76.2} & \textbf{80.4} & 74.8 & \textbf{79.7} & 69.3 & \textbf{83.7} & \textbf{81.5} \\
    \hline
    \end{tabular}
\caption{Task prediction precision for each task ($\%$) at testing time after training the $10$-th task, in comparison with some rehearsal-free methods. The best accuracy scores are in \textbf{bold}.}
\label{tab:taskpred-res}
\end{table*}

\subsection{Experimental Settings}

\paragraph{Datasets.} To evaluate the effectiveness of WAVE-CRE and the baseline models, we utilize two popular datasets:
\begin{itemize}
    \item \textbf{FewRel} \cite{han-etal-2018-fewrel} contains 80 relation types with a total of 56,000 samples. Following the configurations outlined in \citet{wang-etal-2019-sentence}, we split it into 10 non-overlapping sub-datasets. 
    \item \textbf{TACRED} \cite{zhang2017tacred} consists of 42 relations and 106,264 samples.  We adopt the experimental settings proposed by \citet{cui-etal-2021-refining} to partition the dataset into 10 distinct sub-datasets.
\end{itemize}

\paragraph{Baselines.} We compare WAVE-CRE with recent rehearsal-free and prompt-based continual learning methods including L2P \cite{wang2022learning}, HiDe-Prompt \cite{wang2023hide}, and EPI \cite{epi2023}. As these methods were originally designed for computer vision, we re-implemented them for CRE using BERT \cite{devlin2018bert} as the encoder. Additionally, we compare our method with rehearsal-based CRE baselines including EA-EMR \cite{wang-etal-2019-sentence}, RP-CRE \cite{cui-etal-2021-refining}, CRL \cite{zhao-etal-2022-consistent}, CRE-DAS \cite{zhao2023improving}, and CDec+ACA \cite{xia2023enhancing}.

\paragraph{Implementation Details.}
In this work, we used a single NVIDIA A100 for all methods. We tune the hyper-parameters for the proposed model using random search. We maintained a consistent size for the prompt pool $M$ across all tasks. For baselines, we follow the identical experimental settings employed by \citet{zhao-etal-2022-consistent} to ensure fair comparisons. Our proposed model has in total 114M parameters. Since we froze the BERT model, the number of learnable parameters is thus only 3.8M. Training on the FewRel dataset took approximately 7 hours, while for the TACRED dataset, it took approximately 3 hours using our method.


\paragraph{Evaluation Metrics.}  We use the same performance measures (mean accuracy on 5 different random seeds) as in prior work \cite{zhao-etal-2022-consistent} for fair comparison.

\subsection{Main Results}
Table~\ref{tab:fewrel-res} summarizes the performance of all methods on FewRel and TACRED datasets. We begin by comparing WAVE-CRE with rehearsal-free and prompt-based methods for CRE. Notably, among all rehearsal-free methods, WAVE-CRE consistently outperforms across different stages of training on both datasets. Particularly on the last task $\mathcal{T}_{10}$, L2P and EPI exhibit substantially lower performance, with a gap of up to 15\% in final average accuracy compared to our method. This substantial difference underscores the limitations of these existing approaches in addressing catastrophic forgetting across diverse domains. While HiDe-Prompt shows some improvements, it still experiences performance losses of over 15\% and 6\% on FewRel and TACRED, respectively. These losses can be attributed to insufficient task-identity inference techniques, as discussed in Section~\ref{section:detailed_analysis}.

Furthermore, we evaluate WAVE-CRE against recent successful CRE methods, all of which are rehearsal-based. Remarkably, without retaining training data or directly fine-tuning BERT, WAVE-CRE achieves results nearly equivalent to these state-of-the-art baselines on both datasets. Notably, in the FewRel dataset, our method surpasses the latest rehearsal-based methods on the last task. This highlights the significance and effectiveness of our approach.

\subsection{Detailed Analysis} \label{section:detailed_analysis}

\paragraph{Task-specific Prompt Pool.} To illustrate the efficacy of task-specific prompt pools in capturing within-task variations compared to a single prompt approach, we conducted experiments in the task incremental learning setting \cite{vandeven2022three}. In this setting, task identities are provided, ensuring the prompt pool remains identical during training and testing. Table~\ref{tab:ablation_prompt_pool} compares models trained with and without the task-specific prompt pool. Here, ``w/o Prompt Pool" represents using only a single task-specific prompt per task. Our method, WAVE-CRE, trained with the prompt pool, shows a 1.8\% improvement on the final task, demonstrating its effectiveness in enhancing the model's ability to capture within-task variations.


\paragraph{Number of Experts per Prompt $L$.} Similarly, we investigate the effect of different values of $L$ in the task incremental learning scenario. The number of prompts selected, $K$, was chosen to ensure a fair comparison by keeping the total number of experts across experiments equal. The results are presented in Table~\ref{tab:ablation_prompt_length}. As discussed in Section~\ref{section:task-specific-prompt-pool}, setting $L = 1$ allows for flexibility in expert selection and increases the model's expressiveness, as each expert has its own key. This is empirically shown, as our proposal achieves the best performance among different values of $L$.


\paragraph{Detailed Analysis of Task Predictor.} We evaluate our task-ID prediction technique against the baselines EPI and HiDe-Prompt, with results summarized in Table~\ref{tab:taskpred-res}. EPI uses BERT for task identification and Mahalanobis distance to select task-specific parameters, assuming pre-trained representations are well-separated—an unrealistic assumption given their generic origin, leading to poor prediction accuracy. In contrast, our method trains a dedicated task predictor on synthesized query representations, significantly improving accuracy. HiDe-Prompt adopts a similar strategy and improves over EPI but groups all relations within a task as a single class, which can be suboptimal (Section~\ref{section:classifiers}). WAVE-CRE addresses this by using a task predictor with output dimensions corresponding to the number of relations, achieving up to 10\% improvements on both datasets, showcasing our strategy's effectiveness.

\section{Conclusion}
In this work, we propose a novel framework called WAVE-CRE for rehearsal-free continual relation extraction. Our contributions focus on generating representations for replay, precise task prediction, and optimizing within-task and cross-task variability via prompting. These strategies address limitations of current state-of-the-art prompt-based baselines for continual learning. Through extensive benchmarks, we show our model consistently outperforms existing rehearsal-free methods and achieves competitive results with advanced CRE methods. While our methods mitigate catastrophic forgetting, challenges remain. Retaining knowledge of past tasks is difficult, as seen in prior works. Despite improvements using prompt pools for task-specific knowledge, forgetfulness occurs when pools are not properly utilized during testing. Additionally, while prompt pools enhance expressiveness, the current prefix-tuning experts are relatively simple. Future work could explore more complex expert designs to improve the model.


\bibliography{aaai25}

\end{document}